# Low-cost Geometry-based Eye Gaze Detection using Facial Landmarks Generated through Deep Learning

Esther Enhui[*], John Enzhou[*], Joseph Ye[*], Jacob Ye[*], Runzhou Ye[*]

University of Sherbrooke, Sherbrooke, Canada

* All authors contributed equally

**Corresponding author:**

Runzhou Ye

University of Sherbrooke, Sherbrooke, Canada

Email: run.zhou.ye@usherbrooke.ca

## Abstract

**Introduction**: In the realm of human-computer interaction and behavioral research, accurate real-time gaze estimation is critical. Traditional methods often rely on expensive equipment or large datasets, which are impractical in many scenarios. This paper introduces a novel, geometry-based approach to address these challenges, utilizing consumer-grade hardware for broader applicability.

**Methods**: We leverage novel face landmark detection neural networks capable of fast inference on consumer-grade chips to generate accurate and stable 3D landmarks of the face and iris. From these, we derive a small set of geometry-based descriptors, forming an 8-dimensional manifold representing the eye and head movements. These descriptors are then used to formulate linear equations for predicting eye-gaze direction.

**Results**: Our approach demonstrates the ability to predict gaze with an angular error of less than 1.9 degrees, rivaling state-of-the-art systems while operating in real-time and requiring negligible computational resources.

**Conclusion**: The developed method marks a significant step forward in gaze estimation technology, offering a highly accurate, efficient, and accessible alternative to traditional systems. It opens up new possibilities for real-time applications in diverse fields, from gaming to psychological research.

**Introduction**

Gaze estimation, a critical component in understanding human-computer interaction, has seen substantial advancements through the integration of machine learning and computer vision. accurate and real-time gaze estimation is crucial to applications ranging from behavior research [1] to advanced human computer interaction [2-4]. However, the most accurate gaze detection systems (such as the Tobii or Eye Tribe systems) usually require the use of expensive and specialized equipment not applicable to most data collection scenarios.

As reviewed in [5], existing gaze estimation algorithms typically fall into the 3D [6, 7] or 2D model recovery based methods [8-14] and the appearance based methods. The former methods typically require sophisticated devices to produce the outline of the pupil and limbus of the eye and are sensitive to variations in lighting conditions. With advancements in the application of deep learning in a wide array of tasks requiring computer vision ranging from face and pose detection [15, 16] to semantic segmentation [17], the latter methods can be subdivided into feature reduction methods [18] and neural network based methods [19-25]. However, although they are more impervious to variations in lighting conditions and facial morphologies, neural network-based methods typically require large datasets for model training and yield models that are computationally expensive to run in real-time settings.

In this work, we take advantage of the generalizability of novel face landmark detection neural networks that have fast inference speed on consumer-grade chips to generate accurate and stable real-time 3D landmarks of the face and iris. We then derived a small set of geometry-based descriptors of the motion and orientation of the eyes and the head. This set of descriptors/parameters forms an 8-dimentional continuous manifold that describes the x and y direction of eye-gaze. We can then sample points on the manifold and yield linear equations that accurately predict eye-gaze. Herein, we show that it is possible to achieve state of the art estimation accuracy of less than 1.9 degree angular error using this method with negligible computational cost.

# Methods

## Data collection

The primary objective was to capture a series of images based on user-defined events—specifically, mouse clicks. This approach allows for a more controlled and user-driven collection of visual data. Each left-button press by the user triggers the capture of a single frame from the webcam. This process not only records the visual data but also associates it with the specific moment and context of the user's interaction.

Participants interact with the system via a standard mouse. Each left-click represents a significant event or decision by the user, prompting the system to capture an image. Three hundred frames were captured in three separate sessions with five subjects aged from 11 to 26 years old. Subjects were instructed to focus the gaze on the tip of mouse cursor while clicking 20 times on the computer screen at random locations.

## Facial landmark detection

Facial landmarks were detected using the Attention Mesh neural network [26] with the Mediapipe library [15]. The model prediction produces a 3D face mask with 468 landmarks (see **Figure 1**). Each landmark is described by its normalized x, y and z position relative to the center of the image. This model was chosen for its real-time inference speed on consumer electronics and the higher accuracy of landmark detection around the eyes and irises.

## Selection of facial landmarks for gaze inference

Since most of facial landmarks will be intercorrelated as their relative position should not change with changing gaze, only a few landmarks that are independent and relevant to the gaze inference task were selected.

The positions of the medial canthal angle (MCA) of the left ($P_{lMCA}$, landmark_362) and right ($P_{rMCA}$, landmark_133) eyes were used to assess the width of the face in relation to the image frame — as their relative positions to the skull usually remain stable — and the rotation of the head around the y-axis ($R_Y$).

The midpoint between the eyes ($P_{ME}$, landmark_168) and of the bottom of the nose ($P_{BN}$, landmark_2) were used to assess the height of the face in relation to the image frame and the rotation of the head around the x-axis ($R_X$).

The position of the pupil ($P_P$) relative to the face was computed as the center of the top, bottom, left, and right corners of the limbus (landmarks 469, 470, 471, 472 and landmarks 474, 475, 476, 477 for the left and right eyes, respectively).

**Calculation of parameters of eye and head position and orientation**

Since the location of the gaze on the screen depends on the location of the pupil but also on rotations and translations of the head, we computed parameters that capture these motions as follows:

- Rotation of the head around the y-axis ($R_Y$) for left and right gaze (since this angle is typically less than 10 degrees for computer screen viewing, small angle approximation can be used for $tan^{-1}$):

$$R_Y = \frac{P_{lMCA,z} - P_{rMCA,z}}{P_{lMCA,x} - P_{rMCA,x}}$$

- Rotation of the head around the x-axis ($R_X$) for up and down gaze:

$$R_X = \frac{P_{ME,z} - P_{BN,z}}{P_{ME,y} - P_{BN,y}}$$

- Relative height ($H_f$) and width ($W_f$) of the face to the image frame (which are inversely proportional to the distance between the face and the camera):

$$H_f = \sqrt{(P_{ME,x} - P_{BN,x})^2 + (P_{ME,y} - P_{BN,y})^2 + (P_{ME,z} - P_{BN,z})^2}$$

$$W_f = \sqrt{(P_{lMCA,x} - P_{rMCA,x})^2 + (P_{lMCA,y} - P_{rMCA,y})^2 + (P_{lMCA,z} - P_{rMCA,z})^2}$$

- The x and y components of the midpoint between the eyes ($P_{ME,x}$ and $P_{ME,y}$) were used as the x and y position of the head.

- Position of the left ($P_{lP}$) and right ($P_{rP}$) pupils normalized to the relative size of the face:

$$P_{P(x,y)} = P_{lP(x,y)} + P_{rP(x,y)}$$

$$P_{lP(x,y)} = \frac{1}{W_f}\left(P_{lMCA,x} - \frac{1}{4}\sum_{i=469}^{472} Landmark_{i,x}\right), \frac{1}{H_f}\left(P_{lMCA,y} - \frac{1}{4}\sum_{i=469}^{472} Landmark_{i,y}\right)$$

$$P_{rP(x,y)} = \frac{1}{W_f}\left(P_{rMCA,x} - \frac{1}{4}\sum_{i=474}^{477} Landmark_{i,x}\right), \frac{1}{H_f}\left(P_{rMCA,y} - \frac{1}{4}\sum_{i=474}^{477} Landmark_{i,y}\right)$$

**Model construction for gaze prediction**

To assess the form of the model with the above variables, we collected datapoints while the subject rotates the head around the x- and y-axes while keeping the gaze fixed. Since $P_{P(x,y)}$ and

$R_Y$ and $R_X$ behave mostly in a linear fashion, we used multiple linear regression to predict gaze location ($G_{pred}$) on the screen.

$$\begin{bmatrix} G_{pred,x} \\ G_{pred,y} \end{bmatrix} = \begin{bmatrix} [\beta_{0,x} \quad \beta_{1,x} \quad \beta_{2,x} \quad \beta_{3,x} \quad \beta_{4,x}] \times \begin{bmatrix} 1 \\ R_Y \\ P_{P,x} \\ W_f \\ P_{ME,x} \end{bmatrix} \\ [\beta_{0,y} \quad \beta_{1,y} \quad \beta_{2,y} \quad \beta_{3,y} \quad \beta_{4,y}] \times \begin{bmatrix} 1 \\ R_X \\ P_{P,y} \\ H_f \\ P_{ME,y} \end{bmatrix} \end{bmatrix}$$

We fitted data from two recording sessions to the linear equations and used data from the third session to assess model accuracy. We computed the $R^2$ coefficient, the pixel error, and the angular difference between the ground truth gaze ($G_{True}$) and the predicted gaze ($G_{pred}$).

**Statistical analysis and code availability**

Analyses and computations were performed in Python 3.11 with the OpenCV, Mediapipe, and Numpy libraries. All codes written for the present study are available at GitHub: https://github.com/runzhouye.

## Results

**Overview of the gaze inference pipeline**

As shown in **Fig.2**, images of the face were first captured with computer webcam. The Attention Mesh neural network was then employed to detect the location of 478 facial landmarks in real-time. A few relevant landmarks were subsequently selected to compute 8 parameters that describing the geometry of the head and eyes in space. Calibration data were then fitted to multiple linear equations to yield models that predict the location of the gaze on the computer screen.

**Unsupervised dimension reduction with uniform manifold approximation and projection of our eye and head geometry parameters**

We performed unsupervised dimension reduction of our geometric parameters describing eye and head position and orientation using uniform manifold approximation and projection (UMAP) [27].

As shown in **Fig.3**, the set of geometry parameters that we previously constructed naturally ordered each individual image of the face according to the eye-gaze both in the x and the y directions without any supervision. Furthermore, for each subject, unsupervised UMAP ordered each image according to both axes such that the gradient of the x and y position of eye-gaze are mostly perpendicular to each other.

**Accuracy of gaze inference with and without attention mechanism**

The $R^2$ correlation coefficient between the location of eye-gaze was higher in the x-axis (0.98) compared to the y-axis (0.87, $P < 0.003$, **Fig.4A**). The angular error of our gaze inference models was 1.60 degree in the x-axis and 1.95 degree in the y-axis (**Fig.4B**). No significant difference in inference accuracy was observed with *vs.* without the use of attention mechanism in the landmark detection step by the neural network. The average on-screen distance error was lower than 1.5 cm for both the x and y direction with and without attention mechanism (**Table 1**).

As shown in **Fig.5**, our geometry-based regression model accurately predicts the location of the eye gaze on the computer screen in the x- (**Fig.5A**) and y-axis (**Fig.5B**) for all the study participants in the testing dataset.

**Discussion**

In the present work, we present a geometry-based eye-gaze prediction method using facial landmarks obtained with Attention Mesh to achieve state-of-the-art accuracy (under 1.8 degrees of angle error) using only consumer-grade monocular camera. The facial landmark detection neural network that we employed [26] has been designed for high-speed performance, achieving up to 16.6ms processing time on consumer devices. Since the all the computation that our method adds consist of 64 addition/multiplication operations, this should only result in a negligible increase to the total processing cost.

With the advancements in deep neural networks, these have been increasingly applied to computer vision tasks including semantic segmentation [17, 28, 29] and pose estimation [16]. Compared to previous works [30-35], our proposed method has the advantage of not requiring training new convolutional neural networks, which allows it to be more easily implemented.

Dimension reduction techniques such as PCA, t-SNE, and UMAP have been extensively used in genomics studies [36-38]. Unlike previous works that map [18, 39] the entire image of the eye in to very high dimensional manifolds and map them back to a 2 dimensional plane, we derived 8 parameters that describes all the possible movements and orientations of the eyes and the head. Here, using UMAP analysis, we showed that the set of geometry variables that we formulated formed an 8-dimentional manifold that effectively captured the directionality of gaze in both the x and y axes since unsupervised dimension reduction of our variable set resulted in ordering of the images according to gaze direction.

We found that our method resulted in lower prediction accuracy for gaze changes in the y-axis. This can be due to the fact that the left and right sides of the limbus are usually visible, allowing for more accurate estimates of the pupil position on the x-axis, whereas the top and bottom of the limbus are usually occluded by the eyelids, making estimation of the pupil position on the y-axis more challenging, especially for individuals with dark irises. Future works could thus build upon previous reports on detection of the contour of the limbus [14] in order to more accurately determine the elevation of the pupil.

**Conclusion**

This work presents a geometry-based eye-gaze prediction method, achieving state-of-the-art accuracy (less than 1.9-degree angular error) using only consumer-grade monocular cameras. By utilizing facial landmarks detected by the Attention Mesh neural network, we formulated an 8-dimensional manifold of eye and head movements to predict gaze direction accurately and efficiently. Our method's high accuracy, low computational cost, and independence from extensive training datasets mark a significant advancement in gaze estimation technology. The method's overall performance indicates its potential for widespread application in various fields, including human-computer interaction and behavioral research. Future work will focus on improving y-axis prediction accuracy and exploring further applications of this efficient and accurate gaze estimation approach.

**Declaration of Competing Interest**

The authors declare no competing interests.

**Table 1:** Accuracy of gaze inference with and without attention mechanism

|  | Accuracy in the x-axis | | Accuracy in the y-axis | |
|---|---|---|---|---|
|  | Angular error (degree ± SEM) | Distance error (cm ± SEM) | Angular error (degree ± SEM) | Distance error (cm ± SEM) |
| No attention | 1.450 ± 0.112 | 1.04 ± 0.08 | 2.026 ± 0.203 | 1.46 ± 0.15 |
| With attention | 1.751 ± 0.241 | 1.26 ± 0.17 | 1.877 ± 0.144 | 1.35 ± 0.10 |

**Figure legends**

**Figure 1**: Facial landmarks for gaze inference

**Figure 2**: Overview of the gaze inference pipeline

**Figure 3**: Unsupervised dimension reduction with uniform manifold approximation and projection of our geometry parameters. Plots are for each test subjects, color scales represent the ground truth x (**A**) and y (**B**) coordinates of eye gaze.

**Figure 4**: Accuracy of gaze inference with and without attention mechanism, assessed by the R2 correlation coefficient (**A**) and angular error (**B**).

**Figure 5**: Accuracy of gaze inference in the x- and y-axis in the testing dataset. Pixel-wise correlation between the predicted and ground truth locations of eye gaze on the x (**A**) and y (**B**) axes.

**References**

1. Rayner, K., *Eye movements in reading and information processing: 20 years of research.* Psychological bulletin, 1998. **124**(3): p. 372.

2. Zhang, X., Y. Sugano, and A. Bulling. *Evaluation of appearance-based methods and implications for gaze-based applications*. in *Proceedings of the 2019 CHI conference on human factors in computing systems*. 2019.

3. Wang, H., et al. *Hybrid gaze/EEG brain computer interface for robot arm control on a pick and place task*. in *2015 37th Annual International Conference of the IEEE Engineering in Medicine and Biology Society (EMBC)*. 2015. IEEE.

4. Jacob, R.J. and K.S. Karn, *Eye tracking in human-computer interaction and usability research: Ready to deliver the promises*, in *The mind's eye*. 2003, Elsevier. p. 573-605.

5. Cheng, Y., et al., *Appearance-based gaze estimation with deep learning: A review and benchmark.* arXiv preprint arXiv:2104.12668, 2021.

6. Morimoto, C.H. and M.R. Mimica, *Eye gaze tracking techniques for interactive applications.* Computer vision and image understanding, 2005. **98**(1): p. 4-24.

7. Ji, Q. and X. Yang, *Real-time eye, gaze, and face pose tracking for monitoring driver vigilance.* Real-time imaging, 2002. **8**(5): p. 357-377.

8. Yoo, D.H. and M.J. Chung, *A novel non-intrusive eye gaze estimation using cross-ratio under large head motion.* Computer Vision and Image Understanding, 2005. **98**(1): p. 25-51.

9. Zhu, Z. and Q. Ji. *Eye gaze tracking under natural head movements*. in *2005 IEEE Computer Society Conference on Computer Vision and Pattern Recognition (CVPR'05)*. 2005. IEEE.

10. Valenti, R., N. Sebe, and T. Gevers, *Combining head pose and eye location information for gaze estimation.* IEEE Transactions on Image Processing, 2011. **21**(2): p. 802-815.

11. Ishikawa, T., *Passive driver gaze tracking with active appearance models.* 2004.

12. Guestrin, E.D. and M. Eizenman, *General theory of remote gaze estimation using the pupil center and corneal reflections.* IEEE Transactions on biomedical engineering, 2006. **53**(6): p. 1124-1133.

13. Zhu, Z. and Q. Ji, *Novel eye gaze tracking techniques under natural head movement.* IEEE Transactions on biomedical engineering, 2007. **54**(12): p. 2246-2260.

14. Alberto Funes Mora, K. and J.-M. Odobez. *Geometric generative gaze estimation (g3e) for remote rgb-d cameras*. in *Proceedings of the IEEE Conference on Computer Vision and Pattern Recognition*. 2014.

15. Lugaresi, C., et al., *Mediapipe: A framework for building perception pipelines.* arXiv preprint arXiv:1906.08172, 2019.

16. Ye, R.Z., et al., *Effects of Image Quality on the Accuracy Human Pose Estimation and Detection of Eye Lid Opening/Closing Using Openpose and DLib.* J Imaging, 2022. **8**(12).

17. Ye, R.Z., et al., *DeepImageTranslator: A free, user-friendly graphical interface for image translation using deep-learning and its applications in 3D CT image analysis.* SLAS Technol, 2022. **27**(1): p. 76-84.

18. Lu, F., et al., *Adaptive linear regression for appearance-based gaze estimation.* IEEE transactions on pattern analysis and machine intelligence, 2014. **36**(10): p. 2033-2046.

19. Zhang, X., et al. *Eth-xgaze: A large scale dataset for gaze estimation under extreme head pose and gaze variation*. in *Computer Vision–ECCV 2020: 16th European Conference, Glasgow, UK, August 23–28, 2020, Proceedings, Part V 16*. 2020. Springer.

20. Bao, Y., et al. *Adaptive feature fusion network for gaze tracking in mobile tablets*. in *2020 25th International Conference on Pattern Recognition (ICPR)*. 2021. IEEE.

21. Cheng, Y., et al. *A coarse-to-fine adaptive network for appearance-based gaze estimation*. in *Proceedings of the AAAI Conference on Artificial Intelligence*. 2020.

22. Yu, Y. and J.-M. Odobez. *Unsupervised representation learning for gaze estimation*. in *Proceedings of the IEEE/CVF Conference on Computer Vision and Pattern Recognition*. 2020.

23. Guo, Z., et al. *Domain adaptation gaze estimation by embedding with prediction consistency*. in *Proceedings of the Asian Conference on Computer Vision*. 2020.

24. Park, S., et al. *Towards end-to-end video-based eye-tracking*. in *Computer Vision–ECCV 2020: 16th European Conference, Glasgow, UK, August 23–28, 2020, Proceedings, Part XII 16*. 2020. Springer.

25. Chen, Z. and B. Shi. *Offset calibration for appearance-based gaze estimation via gaze decomposition*. in *Proceedings of the IEEE/CVF Winter Conference on Applications of Computer Vision*. 2020.

26. Grishchenko, I., et al., *Attention mesh: High-fidelity face mesh prediction in real-time.* arXiv preprint arXiv:2006.10962, 2020.

27. McInnes, L., J. Healy, and J. Melville, *Umap: Uniform manifold approximation and projection for dimension reduction.* arXiv preprint arXiv:1802.03426, 2018.

28. Ye, R.Z., et al., *Total Postprandial Hepatic Nonesterified and Dietary Fatty Acid Uptake Is Increased and Insufficiently Curbed by Adipose Tissue Fatty Acid Trapping in Prediabetes With Overweight.* Diabetes, 2022. **71**(9): p. 1891-1901.

29. Ye, E.Z., et al., *DeepImageTranslator V2: analysis of multimodal medical images using semantic segmentation maps generated through deep learning.* bioRxiv, 2021: p. 2021.10. 12.464160.

30. Leblond-Menard, C. and S. Achiche, *Non-Intrusive Real Time Eye Tracking Using Facial Alignment for Assistive Technologies.* IEEE Trans Neural Syst Rehabil Eng, 2023. **PP**.

31. Valliappan, N., et al., *Accelerating eye movement research via accurate and affordable smartphone eye tracking.* Nat Commun, 2020. **11**(1): p. 4553.

32. Krafka, K., et al. *Eye tracking for everyone*. in *Proceedings of the IEEE conference on computer vision and pattern recognition*. 2016.

33. Sugano, Y., Y. Matsushita, and Y. Sato. *Learning-by-synthesis for appearance-based 3d gaze estimation*. in *Proceedings of the IEEE conference on computer vision and pattern recognition*. 2014.

34. Funes Mora, K.A., F. Monay, and J.-M. Odobez. *Eyediap: A database for the development and evaluation of gaze estimation algorithms from rgb and rgb-d cameras*. in *Proceedings of the symposium on eye tracking research and applications*. 2014.

35. Zhang, X., et al. *Appearance-based gaze estimation in the wild*. in *Proceedings of the IEEE conference on computer vision and pattern recognition*. 2015.

36. Zhang, Y., et al., *Detection of candidate gene networks involved in resistance to Sclerotinia sclerotiorum in soybean.* J Appl Genet, 2022. **63**(1): p. 1-14.

37. Ye, R.Z., et al., *Adipocyte hypertrophy associates with in vivo postprandial fatty acid metabolism and adipose single-cell transcriptional dynamics.* iScience, 2023.

38. Becht, E., et al., *Dimensionality reduction for visualizing single-cell data using UMAP.* Nat Biotechnol, 2018.

39. Tan, K.-H., D.J. Kriegman, and N. Ahuja. *Appearance-based eye gaze estimation.* in *Sixth IEEE Workshop on Applications of Computer Vision, 2002.(WACV 2002). Proceedings.* 2002. IEEE.

**Figure 1**: Facial landmarks for gaze inference

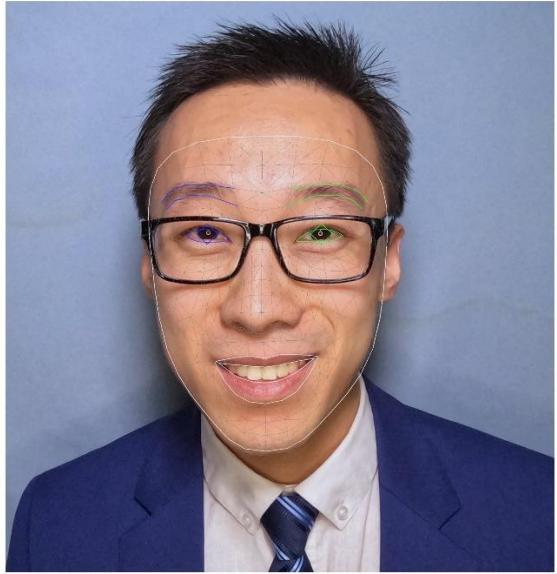

**Figure 2**: Overview of the gaze inference pipeline

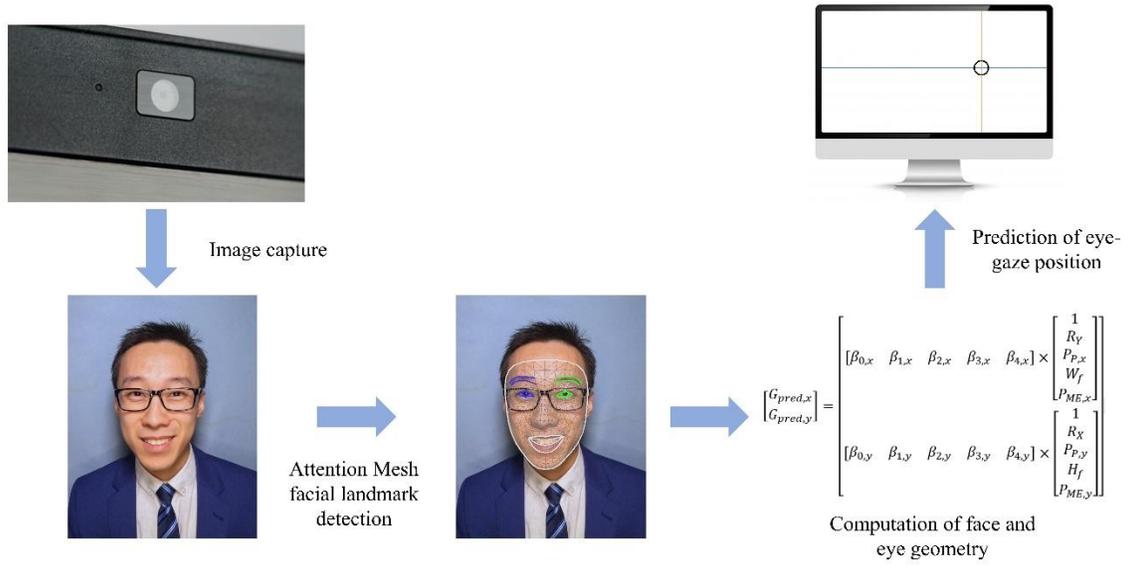

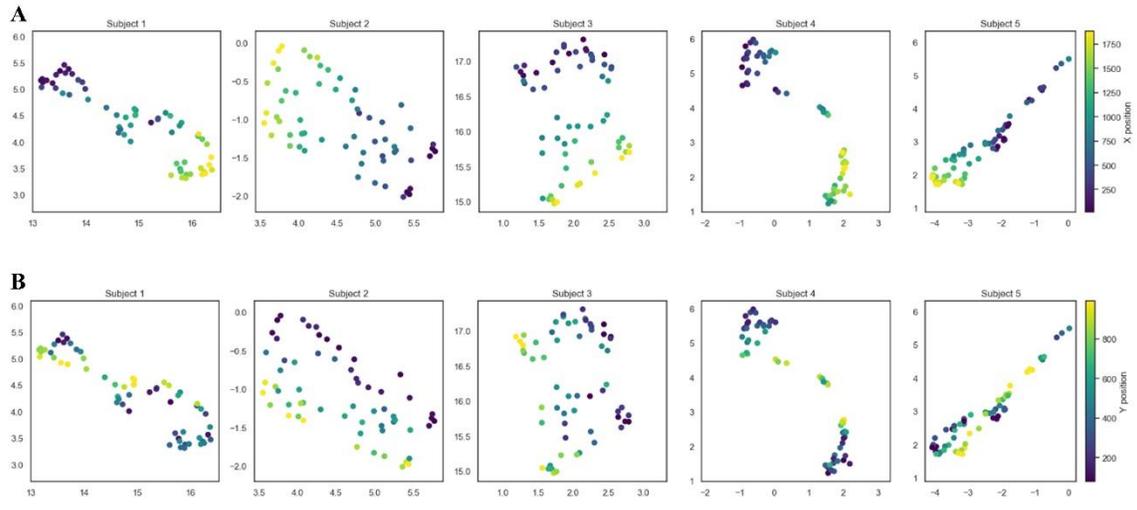

**Figure 3**: Unsupervised dimension reduction with uniform manifold approximation and projection of our geometry parameters

**Figure 4**: Accuracy of gaze inference with and without attention mechanism

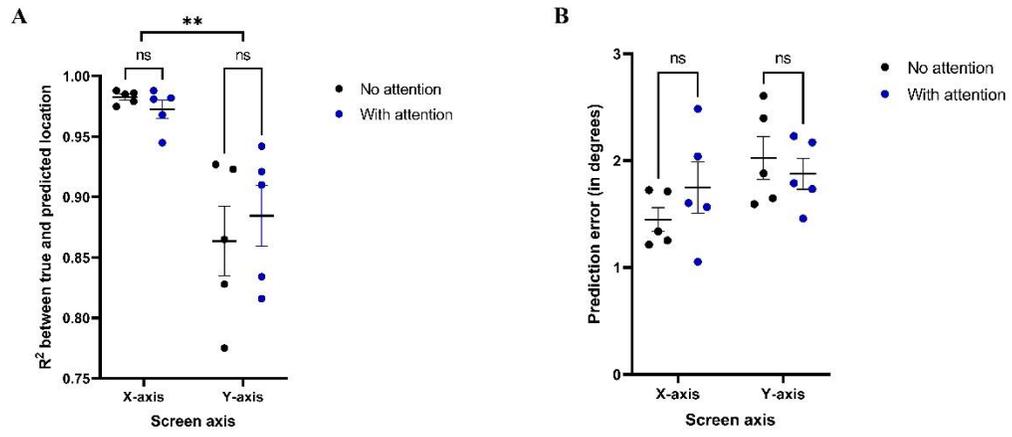

**Figure 5**: Accuracy of gaze inference in the x- and y-axis in the testing dataset

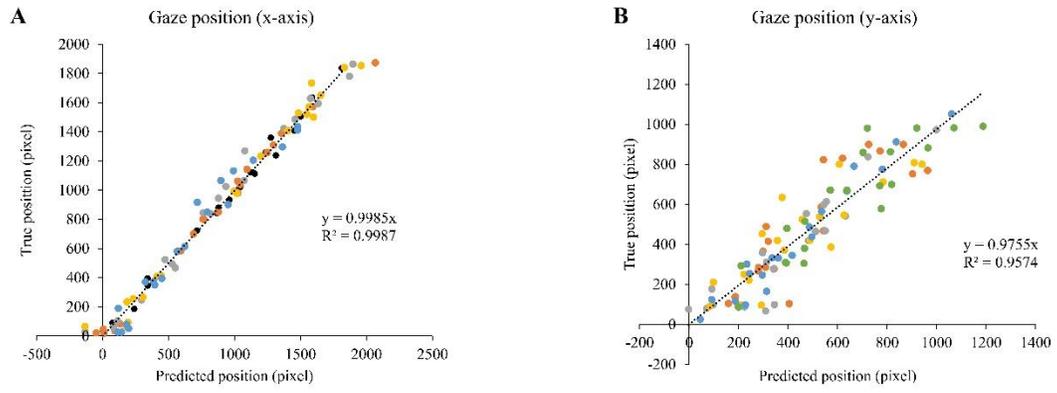